# INDONESIAN-ENGLISH CODE-SWITCHING SPEECH SYNTHESIZER UTILIZING MULTILINGUAL STEN-TTS AND BERT LID

*Ahmad Alfani Handoyo[1], Chung Tran[2], Dessi Puji Lestari[1], Sakriani Sakti[2]*

[1]Bandung Institute of Technology, [2]Nara Institute of Science and Technology

```
13520023@std.stei.itb.ac.id, tran.quang_chung.tq9@naist.ac.jp,
        dessipuji@itb.ac.id, ssakti@is.naist.jp
```

## ABSTRACT

Multilingual text-to-speech systems convert text into speech across multiple languages. In many cases, text sentences may contain segments in different languages, a phenomenon known as code-switching. This is particularly common in Indonesia, especially between Indonesian and English. Despite its significance, no research has yet developed a multilingual TTS system capable of handling code-switching between these two languages. This study addresses Indonesian-English code-switching in STEN-TTS. Key modifications include adding a language identification component to the text-to-phoneme conversion using fine-tuned BERT for per-word language identification, as well as removing language embedding from the base model. Experimental results demonstrate that the code-switching model achieves superior naturalness and improved speech intelligibility compared to the Indonesian and English baseline STEN-TTS models.

## 1. INTRODUCTION

Global connectivity has drastically reduced the time required for message delivery between parties. This heightened connectivity fosters mutual cultural influence among societies, particularly impacting languages. A significant cultural aspect influenced by this connectivity is code-switching—the alternation between two or more languages within a single discourse.

Text-to-Speech (TTS) systems such as FastSpeech 2 [1] and Tacotron 2 [2] traditionally operate within a single language. However, the need for multilingual TTS systems that can handle Indonesian-English code-switching is increasing, driven by sociolinguistic factors like cultural assimilation and the frequent use of English expressions in Indonesian. As a result, code-switching has become prevalent in everyday Indonesian conversations. TTS systems that support code-switching are particularly useful in applications like navigation systems and automated audiobook reading, especially for content that blends both languages.

Despite advancements, existing research lacks a dedicated effort in developing a multilingual TTS system capable of Indonesian-English code-switching. Prior studies have explored multilingual TTS for other language pairs. A study [3] proposed an end-to-end TTS with cross-lingual language models for Mandarin-English, achieving improved naturalness in code-switching speech through shared contextual embeddings across languages. Another study [4] introduced a speech chain framework for semi-supervised learning in Japanese-English TTS, enabling automatic speech recognition (ASR) and TTS to learn from each other with minimal parallel data. Later, the same authors extended this framework to handle multiple language pairs, including zero-shot learning for unseen languages [5].

A major challenge in developing an Indonesian-English code-switching TTS system is the lack of a suitable dataset. Creating such a dataset would require recordings from multilingual speakers fluent in both Indonesian and English. An alternative which is to use mixed datasets from different languages with different speakers would result in inconsistent speech output, with each language sounding as though it is spoken by a different person. To address this, a cross-lingual TTS model that can maintain speaker consistency is needed to leverage existing multilingual datasets and apply them to Indonesian-English code-switching.

STEN-TTS [6] is a non-autoregressive multilingual TTS model based on FastSpeech 2. It has demonstrated effective cross-language adaptation across five languages—English, Mandarin, Japanese, Indonesian, and Vietnamese—and is able to utilize short reference speaker audio. While it does not yet support code-switching, its promising cross-language performance and speaker adaptation capabilities make it a promising foundation for building an Indonesian-English code-switching TTS system.

In this work, we build upon STEN-TTS and integrate a language identification (LID) component using BERT to develop a system capable of synthesizing Indonesian-English code-switching speech. We evaluate the system's performance by analyzing the output speech's intelligibility and naturalness.

## 2. RELATED WORKS

The development of Indonesian TTS systems has evolved significantly over the years. The initial Indonesian TTS was developed by utilizing a diphone concatenation approach with a single speaker, monolingual model [7]. This was followed by the introduction of Hidden Markov Model (HMM)-based which also employed a single speaker and was limited to monolingual capabilities [8]. Recent advancements include neural network-based TTS models such as those using Tacotron 2 for expressive synthesis [9], and generative adversarial network-based systems for improved audio quality [10]. STEN-TTS introduced a multilingual adaptive TTS model, though its application remains predominantly monolingual despite its potential for multilingual adaptation [6]. As observed, most existing Indonesian TTS systems are monolingual and do not address Indonesian-English code-switching.

On the other hand, code-switching in Indonesian-English contexts has been explored, but primarily in ASR systems. For instance, a study proposed rule-based models to improve recognition accuracy [11], while another study used machine speech chain to enhance ASR performance [12]. However, Indonesian-English code-switching in TTS systems remains largely unexplored.

To the best of our knowledge, this study is the first to propose a multi-speaker multilingual TTS system specifically designed to handle Indonesian-English code-switching. This novel approach leverages STEN-TTS to integrate multilingual and adaptive capabilities.

## 3. METHOD

To allow for code-switching between Indonesian and English, a model based on STEN-TTS is proposed as shown in Figure 1.

### 3.1. Monolingual Components in STEN-TTS

While inherently multilingual, STEN-TTS has an internal structure which includes components tailored to specific target languages, limiting output to one language at a time.

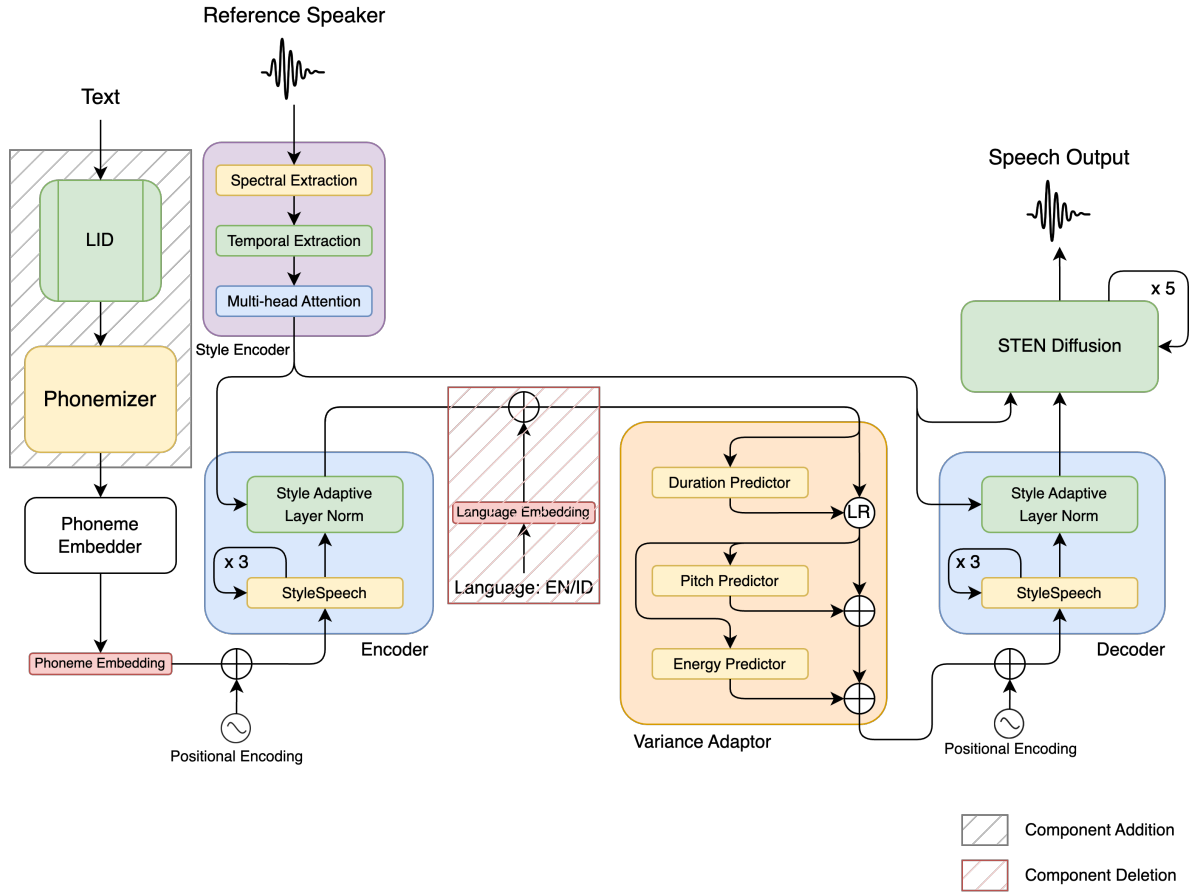

Figure 1: Proposed modification of STEN-TTS to enable code-switching between Indonesian and English. An LID component is integrated before the text-to-phoneme conversion in the Phonemizer, using fine-tuned mBERT. The language embedding, which previously provided language-specific information to the hidden sequence from the encoder, has been removed.

Enabling code-switching between Indonesian and English demands adjustments to these components, enabling the model to recognize and process both languages simultaneously.

Specifically, one identified limiting component is the language embedding, which informs the model about the target language of speech input. Trained on sentence-level text and audio, the language embedding currently characterizes language features at the sentence level but fails to reflect word-level language characteristics crucial for code-switching.

Removing the language embedding component directs the encoder's output directly to the variance adaptor. This approach circumvents the need for retraining the STEN-TTS model, utilizing pre-existing weights. Ultimately, this modification was chosen to handle code-switching scenarios.

### 3.2. LID Approach for Phonemizer Invocation

STEN-TTS employs Phonemizer [13] to transform text from grapheme to phoneme form, essential for TTS input in phoneme embedding format. However, STEN-TTS's current configuration restricts Phonemizer calls to predefined languages among English, Mandarin, Japanese, Indonesian, and Vietnamese.

Challenges arise when the model encounters code-switching, such as between Indonesian and English. In such scenarios, words in code-switched sentences are converted to phoneme sequence according to the model's configured language, disregarding the actual language of each word. For instance, if the model is set to Indonesian, English words are converted to phonemes based on Indonesian phoneme rules, resulting in incorrect pronunciation and stress.

To address this, modifications to Phonemizer invocation are necessary to convert each word to phonemes according to its actual language. A solution involves integrating a Language Identification (LID) component capable of classifying the language of each word before passing it to the Phonemizer for phoneme sequence generation.

However, existing LID models like FastText, trained on a diverse set of 176 languages [14], may misclassify words as languages other than Indonesian or English. Therefore, a more tailored LID approach is needed—one that accounts for the distinct linguistic characteristics of both Indonesian and English to enhance classification accuracy. Recent studies in multilingual LID development have found that using fine-tuned BERT demonstrate superior performance compared to traditional LID methods [15].

In this study, the LID is developed using fine-tuning with mBERT, trained on 104 languages including Indonesian and English [16]. This choice leverages mBERT's ability to represent tokens from both languages effectively. By fine-tuning mBERT, token representations are fed into a two-class fully connected layer, distinguishing tokens as either Indonesian or English. This approach ensures robust language classification, critical for accurate phonetization in code-switched contexts.

## 4. EXPERIMENTS

### 4.1. Dataset

In this study, we utilized several corpora: Wiki-40B [17], ID-EN Code-Mixed [18], Common Voice [19], and CoVoST 2 [20]. The Indonesian portion of Wiki-40B is referred to as the LID-ID dataset, while the English portion is designated as the LID-EN dataset. The ID-EN Code Mixed corpus is labeled as the LID-CS dataset. Transcriptions from Common Voice version 4 in Indonesian were utilized as the Test-ID dataset, whereas CoVoST 2, which translates Indonesian into English, was designated as the Test-EN dataset. Common Voice version 4 was selected due to its direct alignment with the transcribed segments used in CoVoST 2.

The LID-ID, LID-EN, and LID-CS datasets were utilized for fine-tuning the mBERT based LID. The Test-ID and Test-EN datasets served as the foundation for constructing the code-switching evaluation dataset named TestSet, which was used to evaluate both LID and the overall code-switching model. Test-ID and Test-EN consisted solely of transcription texts without audio segments, as the base STEN-TTS model cannot be trained on new audio data. Therefore, these datasets were employed solely for testing LID and code-switching model performance.

The development of the code-switching evaluation dataset, TestSet, was crucial for assessing the LID and code-switching model performance in this study. TestSet was constructed from Test-ID and Test-EN datasets, involving the construction of seven distinct cases. Each case represented different scenarios of text code-switching between English and Indonesian:

1. EN: Entirely in English.
2. ID: Entirely in Indonesian.
3. ID CS 1 word EN: Code-switched text with mostly Indonesian and one English word.
4. ID CS 2 word EN: Code-switched text with mostly Indonesian and two English words.
5. EN CS 1 word ID: Code-switched text with mostly English and one Indonesian word.
6. EN CS 2 word ID: Code-switched text with mostly English and two Indonesian words.
7. Half-Half: Mixed code-switched text with equal parts Indonesian and English.

The construction of the TestSet dataset was carried out in two stages. The first stage involved labelling the language of each word in the Test-ID and Test-EN datasets. The second stage focused on creating the complete dataset for all seven cases, which included assigning language labels to every word in the text.

## 4.2. Implementation

The experiments in this research were conducted on a machine with CUDA acceleration, running Ubuntu 20.04.6 LTS with an NVIDIA Tesla V100 SXM2 GPU (32GB HBM2). A Miniconda virtual environment was set up with various Python tools and libraries to support the experiments. PyTorch version 1.12.1 was selected due to its compatibility with CUDA version 11.4 that was available on the machine.

The LID was developed by fine-tuning the mBERT Multilingual Cased model [16]. However, the imbalance in row counts among the LID-ID, LID-EN, and especially the LID-CS datasets posed a risk of overfitting to the larger datasets. To address this, undersampling was applied to all three datasets, ensuring proportional row counts in line with the smallest dataset, LID-CS. The LID model was developed using a data row proportion of 5:5:1 for LID-ID:LID-EN:LID-CS, with fine-tuning conducted over 10 epochs.

The texts in the datasets were tokenized into representations understandable by the mBERT model input. To allow the model to identify the language of each token, every token was labeled as either Indonesian or English. Subsequently, mBERT was fine-tuned using these tokens along with their language labels. A two-dimensional fully connected layer was added to mBERT to classify the language of each token in the input text. This fully connected layer was trained from scratch during the fine-tuning process.

Next, the CS-TTS code-switching model was developed, integrating the modified STEN-TTS without the language embedding component and the selected LID system from the previous stage. The development involved modifying STEN-TTS by removing the language embedding component and then integrating this modification with the previously fine-tuned LID system.

## 4.3. Evaluation

This phase aimed to evaluate the developed code-switching text-to-speech model. The models used in the evaluation were:
1. STEN-TTS EN: Baseline STEN-TTS model for English.
2. STEN-TTS ID: Baseline STEN-TTS model for Indonesian.
3. CS-TTS: Developed code-switching TTS model.
4. CS-TTS Topline: Developed code-switching TTS model assuming correct language labels from LID to assess the impact of LID.

The evaluation focused on two key aspects: speech naturalness and intelligibility. Both are subjective measures that required participants to listen to outputs from the code-switching models. The evaluation was structured using survey questionnaires divided into two sections—one assessing naturalness and the other intelligibility. In total, there were 7 survey questionnaires, each completed by 5 respondents, totaling 35 respondents.

Speech naturalness was quantitatively measured using Mean Opinion Score (MOS). Respondents listened to audio segments and rated the naturalness of speech from the four tested models on a scale from 1 to 5. A score of 1 indicated highly unnatural speech, while a score of 5 indicated speech approaching human-like quality. These scores are averaged for each model across each code-switching case. Speech naturalness was also qualitatively assessed through ranking models by respondents. To ensure evaluation objectivity, respondents were not informed of the types of models being tested, and the order of models was randomized.

To measure speech naturalness, the TestSet dataset was used. Seven text samples were selected from TestSet for the 7 code-switching cases pronounced by the four models, resulting in a total of 49 texts tested. Each speech naturalness section in the survey questionnaire contained 7 questions, each for each code-switching case. For each question, respondents rated the speech naturalness of outputs from the four models and ranked them for one code-switching case. With 7 questions and four models, each respondent listened to a total of 28 audio segments. With 7 questionnaires, a total of 196 audio segments were evaluated. The allocation of code-switching cases to survey questions was structured so that each of the 196 audio segments was listened to by exactly 5 respondents in one survey questionnaire.

Speech intelligibility was assessed using the Sentence Understanding Score (SUS) method. Respondents listened to sentences designed to be syntactically sensible but semantically nonsensical. They were then asked to transcribe what they heard. The transcribed text was compared against the intended text to calculate the Word Error Rate (WER), to evaluate the intelligibility of speech produced by the models.

To measure speech intelligibility, a set of 14 SUS sentences containing Indonesian-English code-switching was created. Each speech intelligibility section in the survey questionnaire contained 8 randomized questions, with each of the four models tested by 2 questions. For each question, respondents were asked to transcribe SUS sentence outputs of Indonesian-English code-switching produced by one of the four models. With 14 SUS sentences spoken by 4 models, a total of 56 audio segments were evaluated. The allocation of 14 sentences for 4 models to survey questions was structured so that each of the 56 audio segments was listened to by exactly 5 respondents in one survey questionnaire.

The questionnaires were distributed to 35 students and alumni from the Computer Science and Information Systems and Technology Information Study Programs at the Bandung Institute of Technology (ITB). All respondents were native Indonesian speakers who regularly use English in both professional and academic settings.

## 5. RESULTS AND DISCUSSION

### 5.1. Speech Naturalness

The Mean Opinion Score (MOS) results in Table 1 reveal varying performance among TTS models depending on the language and code-switching scenarios. In monolingual English cases (EN), the baseline STEN-TTS EN model surpasses both the CS-TTS Topline and CS-TTS code-switching models by around 0.4 MOS points. In contrast, for monolingual Indonesian cases (ID), the MOS score of the baseline STEN-TTS ID model is comparable to CS-TTS Topline and CS-TTS, with the STEN-TTS ID model scoring between the two and outperforming CS-TTS. This suggests that in monolingual cases, the developed CS-TTS Topline and CS-TTS models perform comparably to their respective baseline models for each language.

Table 1: MOS Comparison of TTS Models

| Case | TTS Model | | | |
|---|---|---|---|---|
| | STEN-TTS EN | STEN-TTS ID | CS-TTS | CS-TTS Topline |
| EN | 3.571 | 1.114 | 3.171 | 3.114 |
| ID | 1.286 | 3.114 | 2.971 | 3.371 |
| ID CS 1 word EN | 1.257 | 2.400 | 3.714 | 3.600 |
| ID CS 2 word EN | 1.286 | 2.229 | 3.429 | 3.800 |
| EN CS 1 word ID | 3.086 | 1.114 | 3.429 | 3.514 |
| EN CS 2 word ID | 2.571 | 1.343 | 3.429 | 3.686 |
| Half-Half | 2.086 | 1.571 | 3.514 | 3.714 |
| Total | 2.163 | 1.841 | 3.379 | 3.543 |

In unbalanced code-switching cases—ID CS 1 word EN, ID CS 2 word EN, EN CS 1 word ID, and EN CS 2 word ID—there is an improvement in MOS scores for the CS-TTS Topline model followed by CS-TTS, compared to both baseline models. Specifically, in the ID CS 1 word EN case, the CS-TTS model achieves an MOS score of 3.714, which is higher than CS-TTS Topline at 3.6. However, this 0.114 difference is relatively small and negligible.

In balanced code-switching cases like Half-Half, both CS-TTS and CS-TTS Topline models also show increased MOS scores over baseline models, indicating that these models handle language transitions better than baseline models. Overall, CS-TTS and CS-TTS Topline models demonstrate more consistent and superior performance across various language scenarios compared to baseline models.

Referring to the model ranking results in Figure 2, it can be observed that CS-TTS Topline is frequently selected as the top-ranked model in terms of speech naturalness, followed by CS-TTS in second place, with baseline models ranked lower. This qualitative ranking suggests that the developed code-switching models excel compared to baseline models.

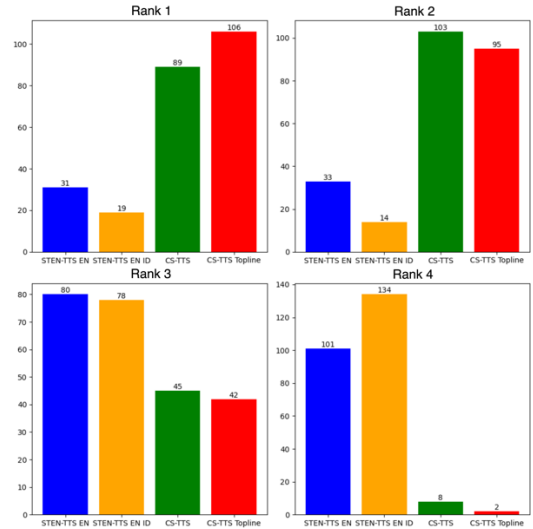

Figure 2: Ranking of TTS models based on preference.

### 5.2. Speech Intelligibility

Table 2: WER Comparison of TTS Models

| Model | WER |
|---|---|
| STEN-TTS EN | 42.18% |
| STEN-TTS ID | 36.44% |
| CS-TTS | 17.43% |
| CS-TTS Topline | 12.87% |

The Word Error Rate (WER) results for SUS sentence transcriptions, presented in Table 2, show that the CS-TTS Topline model achieves the best performance with a WER of 12.87%. It is followed by CS-TTS with a WER of 17.43%, STEN-TTS EN at 42.18%, and STEN-TTS ID at 36.44%. The superior WER of the CS-TTS Topline model compared to CS-TTS can be attributed to the impact of the LID component in CS-TTS, which may misclassify language labels. The improved performance of both the CS-TTS and CS-TTS Topline models over the STEN-TTS EN and STEN-TTS ID baseline models indicates that the code-switching approach significantly reduces sentence pronunciation errors.

## 6. CONCLUSION

The handling of Indonesian-English code-switching in STEN-TTS involved removing the language embedding component and adding an LID component, which identifies the language of the input text on a per-word basis using fine-tuning of BERT. This approach effectively improved speech naturalness, as evidenced by the increase in MOS score of the code-switching model compared to the English and Indonesian baseline STEN-TTS models. Additionally, the code-switching model demonstrated better speech intelligibility, reflected by a decrease in WER on SUS sentences relative to the baseline models.

However, the removal of the language embedding component led to variations in phoneme pronunciation and longer durations in audio, including increased lengths of phonemes and pauses between words. This suggests that while this adjustment improved naturalness and intelligibility, it may have inadvertently impacted the timing and rhythm of the synthesized speech. Moreover, the model encountered errors in pronouncing the vowel "e" in Indonesian words due to inaccuracies in the phoneme sequence generated by Phonemizer.

## 7. ACKNOWLEDGMENT

Part of this work is supported by JSPS KAKENHI Grant Numbers JP21H05054 and JP23K21681, as well as JST Sakura Science Program.